\documentclass[letterpaper, 10 pt, conference]{ieeeconf}  

\IEEEoverridecommandlockouts                              
\usepackage{graphics} 
\usepackage{epsfig} 

\usepackage{amsmath} 
\usepackage{amssymb}  
\usepackage{subcaption} 
\usepackage{algorithm}
\usepackage[noend]{algpseudocode}
\usepackage{siunitx}
\usepackage{textcomp}
\usepackage{gensymb}
\usepackage{csquotes}

\usepackage[hidelinks]{hyperref}
\hypersetup{
  colorlinks   = true, 
  urlcolor     = blue, 
  linkcolor    = blue, 
  citecolor   = red 
}

\usepackage{color}

\title{\LARGE \bf
Hybrid Multi-camera Visual Servoing to Moving Target
}

\author{Hanz Cuevas-Velasquez$^{1\dagger}$, Nanbo Li$^{1\dagger}$, Radim Tylecek$^{1}$, Marcelo Saval-Calvo$^{2}$ and Robert B. Fisher$^{1}$
\thanks{*Authors acknowledge the support of EU project TrimBot2020 and a project from University of Alicante (Gre16-28).}
\thanks{$^{1}$: School of Informatics, University of Edinburgh, contact:
{\tt\small rtylecek@inf.ed.ac.uk}; $^{2}$: Dept. Computer Technology, University of Alicante; 
        $^\dagger$: joint first authors}%
}

\begin{document}

\maketitle
\thispagestyle{empty}
\pagestyle{empty}

\begin{abstract}

Visual servoing is a well-known task in robotics. However, there are still challenges when multiple visual sources are combined to accurately guide the robot or occlusions appear. In this paper we present a novel visual servoing approach using hybrid multi-camera input data to lead a robot arm accurately to dynamically moving target points in the presence of partial occlusions. The approach uses four RGBD sensors as Eye-to-Hand (EtoH) visual input, and an arm-mounted stereo camera as Eye-in-Hand (EinH). A Master supervisor task selects between using the EtoH or the EinH, depending on the distance between the robot and target. The Master also selects the subset of EtoH cameras that best perceive the target. When the EinH sensor is used, if the target becomes occluded or goes out of the sensor's view-frustum, the Master switches back to the EtoH sensors to re-track the object. Using this adaptive visual input data, the robot is then controlled using an iterative planner that uses position, orientation and joint configuration to estimate the trajectory. Since the target is dynamic, this trajectory is updated every time-step. Experiments show good performance in four different situations:
tracking a ball, targeting a bulls-eye, guiding a straw to a mouth and delivering an item to a moving hand. The experiments cover both simple situations such as a ball that is mostly visible from all cameras, and more complex situations such as the mouth which is partially occluded from some of the sensors.
\end{abstract}

\section{INTRODUCTION}
The range of robotic applications has greatly increased with the advent of  low-cost 3D sensing technology. Among the different new uses of robots, social interaction is one of the more exciting areas of research and development. But these applications require methods to guide robots to perform tasks that interact with humans, {\it e.g.} emptying a spoon into a mouth, offering tools, pouring liquids for people, etc. One factor that these tasks have in common is the motion of the target, which motivates in part the research presented here.

Visual servoing methods, iteratively and in real-time, control robots using visual information as input data. There is much previous research into visual servoing and good surveys exist \cite{Janabi-Sharifi2011, Kazemi2010, Finn2017}, including a recent  survey of medical robotics servoing applications \cite{Azizian2014}. 
To control the robot, cameras can be placed on the robot arm (eye-in-hand) or in the environment (eye-to-hand). These terms have been defined as:  \textit{``the camera is said eye-in-hand (EinH) when rigidly mounted on the robot end-effector and it is said eye-to-hand (EtoH) when it observes the robot within its work space''} \cite{Flandin2000}. 
Our hypothesis is that using a hybrid scheme we can switch to the best sensor (EinH $\Leftrightarrow$ EoH) in terms of accuracy, which is typically the EinH camera in close range from the target.
Impressive results have been reached using eye-to-hand cameras, such as the catching flying objects \cite{Bauml2010, Bauml2011, Kim2014}. The system learns how to catch objects using several cameras and with a human initially manipulating the arm. Bauml {\it et al.} \cite{Bauml2010, Bauml2011} used a trajectory model so that the ball movement and catch position could be predicted. However, in \cite{Kim2014} statistically and dynamically unbalanced objects (half full bottles or a racket) are used, hence they readjust the near future predicted target position iteratively. Other approaches solve occlusion problems using  using multiple cameras in the environment, such as the work of Maniatis {\it et al.} \cite{Maniatis2017} where they fuse multiple RGBD sensors around the arm, creating an occupancy space to find empty areas where a robot-mounted camera could be placed.

Multi-camera setups that combine data from external and arm-mounted sensors \cite{Lippiello2005,Kermorgant2011,Bdiwi2017} acquire information from different perspectives to solve problems such as occlusion, high precision targeting via coarse-to-fine positioning, dynamic target acquisition, etc. 
When multiple sensors are used, the configuration could be eyes-in-hand along with eyes-to-hand.
Quintero {\it et al.} \cite{Quintero2014} explored both EinH and EtoH, using stereo sensors in hand but not as a 3D sensor and used RGB data separately. 
Wang {\it et al.} \cite{wang2013} servoed to dynamic targets in cases where the data capture is slower than the target motion. They use visual sensing dynamics to compensate for the slow sampling and large latency of the visual feedback. 
Hybrid EinH/EtoH was used in various approaches. Lippiello et al. presented in \cite{Lippiello2007} an approach where all sensors are included in the pose estimation model. 
On the other hand, Chang and Shao \cite{Chang2010} used  EtoH (RGB camera) to coarsely locate the target pose, and EinH (laser projector and a camera) to control the fine position of the robot moving towards the target.

In the research reviewed above, the image data is analyzed using traditional algorithmic methods. However, some research approaches are analyzing the the visual information using deep network methods. Lee \textit{et al.} \cite{Lee2017} used deep features to learn a visual servo mapping from image to motor control, in a manner more robust to visual variation, changes in viewing angle and appearance, and occlusions. 
Zhang \textit{et al.} \cite{Zhang2015} trained a \textit{Deep~Q Network} to servo based on simulation, using image data inputs. 

There have been many approaches to visual servoing, including EinH, EtoH and hybrid schemes. However, there are still some challenging problems like perception of large scenarios with multiple EtoH, or avoiding self-occlusions with the robot and the visual system. 
To cope with such problems, this paper presents a novel approach for visual servoing using a hybrid-camera setup that combines a 3D EinH and multiple-EtoH 3D sensors for dynamic targets. The method uses a Master process that selects the input information for the servoing from a global 3D EtoH virtual sensor or a 3D stereo EinH sensor, depending on the distance to the target and perception quality. Global scene analysis uses 3D data fused from multiple RGB-D sensors, where only those with good quality perception are selected for fusion. If the target is close enough, the EinH sensor is used for control; otherwise, or if the target moves out the view of the EinH stereo 3D sensor, the whole set of EtoH sensors are activated. This solution allows a better visualization of objects and to overcome partially covered targets. 
The main contributions of the paper are:
\begin{enumerate}
\item {\bf A novel robot workcell} incorporating multiple RGBD sensors, an inverted robot and an arm-mounted real-time stereo sensor that supports 3D capture and servoing over a range of scales (Section \ref{sec:problem}).
\item {\bf A hybrid 3D servoing algorithm} using data from both the global (for coarse alignment) and arm-mounted (for fine alignment) 3D sensors (Section \ref{sec:proposal}).
\item {\bf A source switching algorithm} that selects between the global and arm-mounted 3D sensors for most accurate performance (Section \ref{sec:proposal}).
\end{enumerate}


\section{Problem Statement} \label{sec:problem}

\begin{figure}[!th]
\includegraphics[width=0.48\textwidth]{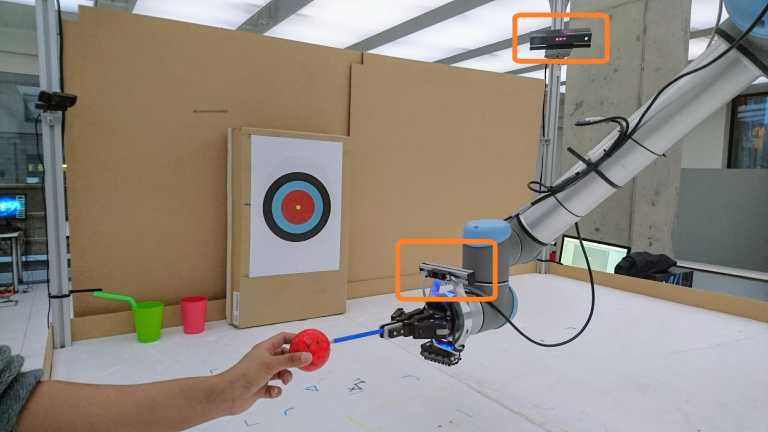}\\

\caption{Setup for tracking of targets with multiple depth sensors. The upper orange square shows one of the 4 RGBD Kinects, and the lower orange square marks the stereo EinH sensor. 
}
\label{fig:setup}
\end{figure}

This paper presents a novel hybrid multi-camera eye-to-hand (EtoH) / eye-in-hand (EinH) approach  to guide a robot arm in different tasks. The target point is assumed to be dynamic, which makes the problem more complex in terms of the switching between EtoH and EinH servoing as the spatial relationship between the robot and target changes. 
The proposed EtoH/EinH switching algorithm is general, but for experimental evaluation the workcell seen in Fig.~\ref{fig:setup} is used, which has these components:
\begin{itemize}
\item Inverted UR10 arm and work surface.
\item Video rate stereo sensor~\cite{Honegger2017} {(720x480 color pixels, 30 selectable depth planes, 10 fps)} mounted on the UR10 arm bracket (see bottom orange square Fig.~\ref{fig:setup}).
{The sensor's view-frustum is 45$^\circ$ wide and bounded between 20 and 40 cm from its mounting point, resulting in approximately 0.7 cm depth quantization.}
\item Four Kinect v2 RGB-D sensors at the four corners of the workcell (Fig.~\ref{fig:camera-data}).
\end{itemize}

\subsection{Proposed Approach}\label{sec:proposal}

\begin{figure}[tbh]
    \includegraphics[width=1\linewidth]{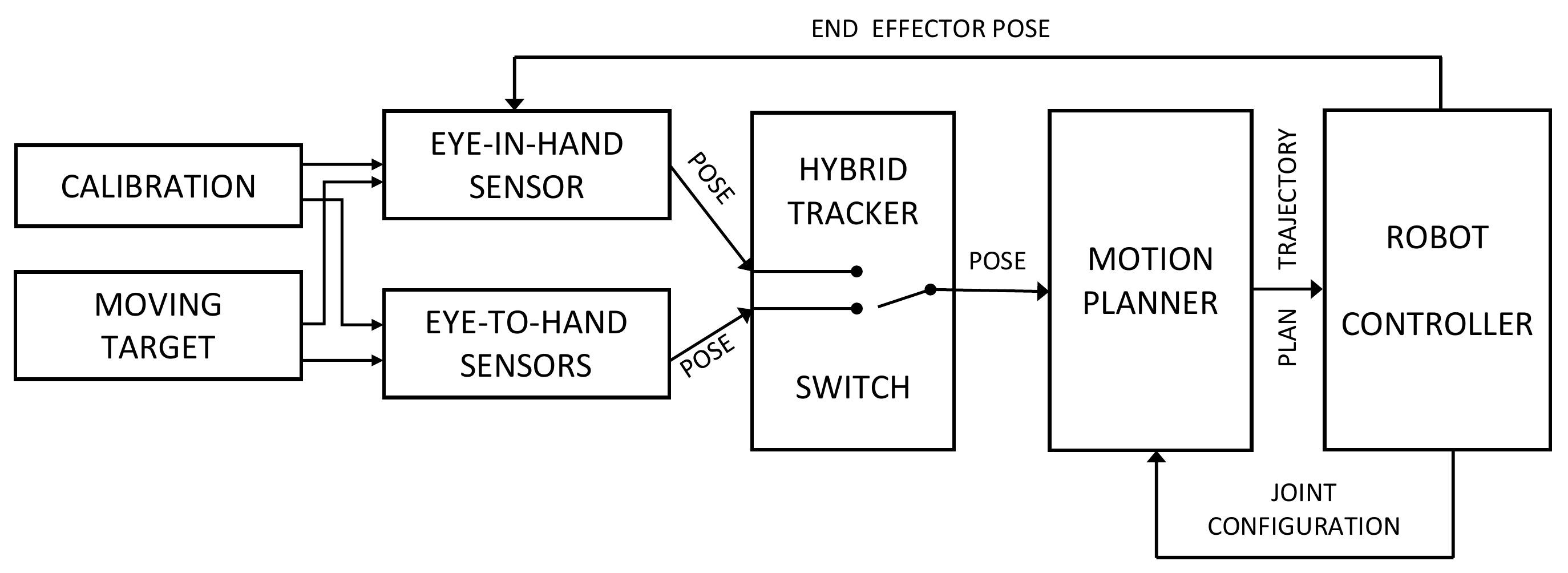} 
	\caption{Components of the position-based visual servo system. Sensor calibration is carried out before the servoing loop is executed, then all components run in parallel. The Master controller switches between the EinH or EtoH input data. Target tracking is performed using the selected data, with target position being given to the motion controller.}
	\label{fig:schematic}
\end{figure}

The proposed visual servoing approach uses a hybrid multi-camera setup, an iterative color 2D target segmentation and a 3D target location algorithm switching between EtoH and EinH sensors to accurately locate the target and thus position the robot arm for a specific purpose.
The system schema (Fig.~\ref{fig:schematic}) shows the main software components, which are discussed in detail below. The implementation uses a combination of ROS and custom Matlab specialist packages.

The image data can come from any or several of the four Kinect RGBD sensors (EtoH), or the arm-mounted stereo sensor (EinH), and as with any position based servoing, their good calibration is critical to the accuracy of the system. 
The intrinsics of the Kinect cameras are calibrated using Kalibr~\cite{Furgale2013}. The extrinsic calibration to register depth data from the four cameras into a common global coordinate system is carried out in two steps. First similar to~\cite{Su2018} a spherical marker is placed in different locations across the workspace and the center of the sphere in each camera a is calculated from segmented point cloud. Next, Procrustes analysis of the corresponding centers is used to find the transformation from each camera to the reference.
Finally, \textit{april-grid} pattern~\cite{Furgale2013} placed on the tabletop provides a transform of the workcell's center and orientation to a reference Kinect, resulting in a fused point-cloud of the whole workcell (Fig.~\ref{fig:point-cloud}). The residual distance of the corresponding marker center points after the registration was $\sim$3 cm on average and increased towards the corners of the workcell.
The EinH stereo sensor is similarly calibrated with respect to the gripper mounted at the end of the robot arm, whose global position can be derived from the current robot configuration. In the static case the combined EinH error of $\sim$1 cm is significantly lower than EtoH, which is the main motivation to use it when possible, leading to the advocated hybrid scheme. 
Based on our initial experiments we adopted the sensor switching strategy to only use sensors close to a target location, which provide less noisy data and more accurate target poses compared to sensors far from the target. 
Our attempts to continuously average detections from all sensors (e.g. using Kalman filter) has led to inferior accuracy and reduced overall system performance (more data bandwidth and processing resources needed).  

The core of the system is a Master state machine that connects the robot-arm control with the image analysis. The Master also decides when to use the EtoH or the EinH sensors. This switch depends, mainly, on whether the target is in the view-frustum of the eye-in-hand sensor. The stereo pair mounted on the arm is meant for fine accuracy, but its working range is narrow and close in distance (see parameters in the component list \ref{sec:problem}). 
Initially the EtoH sensor provides the image data, used to servo the robot towards the moving target. Once the robot is close enough, the Master switches to the EinH stereo sensor. 
If the target goes out of the EinH view-frustum, the master switches back to the EtoH multi-camera to servo the target back into the EinH range.  
{There is a 5 cm hysteresis difference threshold when switching from EtoH to EinH to limit oscillation at the switching boundary. There is no hysteresis when switching from EinH to EtoH.}

Visual servoing uses the input information selected by the Master in two different components: Object tracking (Sec.~\ref{sec:objtrack}), and robot controller with position based control (Sec.~\ref{sec:position}), which are described in more detail below.

\subsection{Hybrid Object Tracking}\label{sec:objtrack}

The visual tracking subsystem combines inputs from multiple RGBD sensors to estimate the moving target's position by optimally selecting active sensors, particularly in cases when the target becomes occluded by the robot arm or the operator, or leaves the view-frustum of the EinH sensor.
Color thresholding in the Lab color space and morphological post-processing gives the target's 2D image position. As target detection is not a main point of this paper, the targets are easily distinguishable (Fig.~\ref{fig:targets}). In the case of circular targets, we neglect the effect of perspective projection and assume the projected shape is approximately circular. 
The detection component could be replaced by a trainable object detector such as~\cite{Redmon2016}. 

The 3D target position $s(t)$ is estimated using the registered point cloud value associated with each RGB image pixel. Color segmentation finds the target's image region which gives an associated set of 3D points, whose center of mass estimates the target 3D position. 
A 3D target position is estimated for each active EtoH sensor and then averaged to get a more precise location of the target (because of errors in the global registration of the four Kinect sensors). Normally the fusion uses only the 3D positions from the two Kinects that are closest to the target and have it in their field of view. The data from all four Kinects is used if target detection fails.

When the object is in the view-frustum  of the EinH camera,  tracking switches from EtoH to EinH (which provides color image and depth disparity). As before, color information is used to segment the target. Then, the center of mass is estimated using the 3D point cloud of the segmented object in the disparity map. This position is in the EinH camera coordinate system, which is then transformed into the global 3D space, by using the current UR10 arm joint angles of the arm to obtain the current camera pose. 


\subsection{Position Based Motion Control} \label{sec:position}
Visual servoing to moving targets requires fast movement control of the robot arm and real-time motion planning. 
To plan motions in the presence of a human operator, safe movements are needed. The kinematic planning uses spatial position constraints and plans motion in joint space with trajectory interpolation for better stability.

\begin{figure}[tb]
 \centering
 \includegraphics[width=1\linewidth]{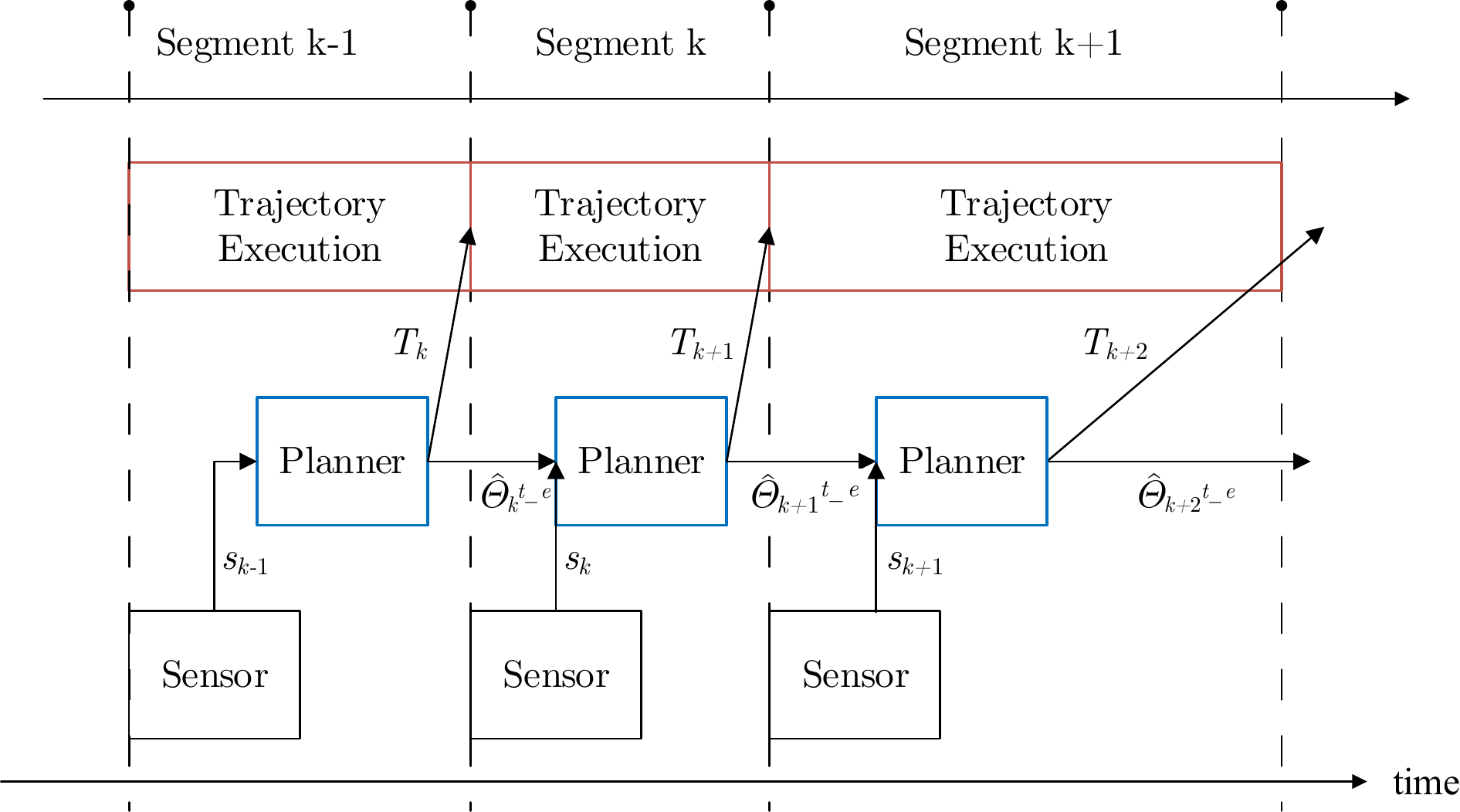}\quad
 \caption{ 
 Parallel replanning and execution scheme. See text for explanation.}
 \label{fig:replan}
\end{figure}

A segment $k$ is a variable time period during which a given plan is executed and simultaneously the next trajectory is planned based on the current sensor reading (Fig.~\ref{fig:replan}). Define $t_0$ and $t_e$ as the starting and ending times of trajectory segment $k$. All variables that change within a segment will be parametrized with $t \in [t_0, t_e]$. The tracked target position obtained from the visual tracker at time $t_0$ in the segment $k$ is $s_k = s(t_0)$.
 
The next end-effector goal pose $\textbf{X}^{*}_{k} = [y^{*}_{k}(t_e), a^{*}_{k}(t_e)]^T$ in task-space for segment $k$ is given to the motion planner in the previous segment $k-1$. From now on, all variables with superscript $\ast$ are target values for time $t_e$. $\textbf{X}^{*}_{k}$ combines the desired position $y^{*}_{k}$ of the robot end-effector and the desired orientation $a^{*}_{k}$. Similarly, the current actual robot end-effector pose is denoted $\textbf{X}_k(t)=\,[y_{k}(t), a_{k}(t)]^T$.

Arbitrary target motions make its next appearance less predictable, {\it i.e.} uncertainty needs to be considered when estimating $y^{*}_{k}$. For this reason, an iterative approach strategy is used (Sec. \ref{sec:planning}). A movement ``discount'' factor $\alpha \in (0, 1]$ (here 0.8) compensates for the unpredictability when calculating $y^{*}_{k}$:
\begin{equation} \label{eq1}
y^{*}_{k} = y_{k}(t_0) + \alpha*(s_{k-1} - y_{k}(t_0)) 
\end{equation}
where $y_k(t_0) = y^{*}_{k-1}(t_e)$ is the initial task-space position of the robot end-effector.

Equation \eqref{eq1} defines the servoing to a target by moving the robot towards the estimated orientation rather than the estimated position. This procedure iteratively leads the robot's end-effector to the target point until convergence. When the target is less than 2 cm away, the discount used is $\alpha = 1.0$.

In joint space, $q_{k}(t) \in \mathbb{R}^6$ is the current joint configuration ($6$ DoF) at time $t$ and $q^{*}_{k}$ is the desired joint configuration at time $t_e$. The robot state is described with $R(t) = [ y_k(t), a_k(t), q_k(t) ]$ and the task cost function $f$ is: 
\begin{equation} \label{eq2}
\begin{split} 
 f(R(t)) &=\|y^{*}_{k} - y_k(t)\|^2_{W1} +\\
&+ \|a^{*}_{k} - a_k(t)\|^2_{W2} + \|q^{*}_{k} - q_k(t)\|^2_{C}
\end{split}
\end{equation}
where $W1 \in \mathbb{R}^{3\times 3} ,\, W2 \in \mathbb{R}^{3\times 3},\,$ and $C \in \mathbb{R}^{6\times 6} $ are empirically set diagonal weight matrices for each criterion. The planned end effector position $y^{*}_{k} = [y^{*}_{kx},\, y^{*}_{ky},\, y^{*}_{kz}]^T$ is constrained to lie in a bounding box given by the workcell dimensions, and the end effector orientation $a^{*}_{k} = [\text{sin}(\gamma^{*}),\text{cos}(\gamma^{*}),0]^T$ is constrained to point towards the side of the workcell where the human operator stands, with angle $\gamma^{*} = \text{yaw}(s_{k-1} - y_k(t_0))$ derived from the relative target location. The actual constraints are:
\begin{align}
  -0.9 &\leq y^{*}_{kx}\leq 0.9, &
  -0.9 &\leq y^{*}_{ky}\leq 0.9, \nonumber \\
   0.2 &\leq y^{*}_{kz}\leq 1.2, &
  -\frac{\pi}{4} &\leq \gamma^{*} \leq \frac{\pi}{4} \nonumber
\end{align}

We use the ROS MoveIt! Cartesian path planner to minimize the objective function \eqref{eq2} on the current segment time period $(t_0,t_e)$
with several joint space waypoints (depending on the distance), obtained by interpolating waypoints between {$q^{*}_{k-1}$ and $q^{*}_{k}$} to increase the smoothness of the trajectory. 
The maximum velocity~$\dot{q}_k(t)$ and acceleration~$\ddot{q}_k(t)$ are limited. 

 The planned trajectory is represented as $T_k = [ \Theta_k^{t\_0},  \Theta_k^{t\_1}, \dots , \Theta_k^{t\_e} ]$, where $\Theta_k^{t\_0} \doteq \Theta_{k-1}^{t\_e}$. Any waypoint state $\Theta_k^t$ within the fine-interpolated trajectory segment $T_k$ has now the desired joint position, velocity and acceleration at time $t$, i.e. $\Theta^t_k=[q^t_k,\, \dot{q}^t_k,\,\ddot{q}^t_k]$. 

\subsection{Planning Strategy}\label{sec:planning}
To implement an iterative servoing process with a moving target, re-planning is necessary to keep the target positions and generated trajectories updated. The planner typically takes about 30ms per segment to generate a new trajectory {which typically takes 300ms to execute}.  Sequentially alternating trajectory planning and execution will not only significantly increase time cost, but also risks a failed approach sometimes due to target motion. Hence, planning and execution proceed in parallel to improve the efficiency. 
{As shown in Fig.~\ref{fig:replan}, the planned trajectory at time segment $k$ is a set of waystates $T_{k+1} = [ \Theta_{k+1}^{t\_0},  \Theta_{k+1}^{t\_1}, \dots , \Theta_{k+1}^{t\_e} ]$, where $\Theta_{k+1}^{t\_0} \doteq \Theta_{k}^{t\_e}$ (because the actual motion will result in a slightly different state). Any waystate $\Theta_{k+1}^t$ within the fine-interpolated trajectory plan $T_k$ should have the desired joint position, velocity and acceleration at time $t$, i.e. $\Theta^t_{k+1}=[q^t_{k+1},\, \dot{q}^t_{k+1},\,\ddot{q}^t_{k+1} ]$. This trajectory is computed given the expected final joint state $\Theta_{k}^{t\_e}$ from the previous segment and the current estimated target pose in cartesian space $X_k(t_{current})$.
As the new trajectory $T_{k+1}$ is planned while the current trajectory is still being executed, the initial pose for segment $k+1$ is approximated by $\Theta_k^{t\_e}$.
}
{A segment finishes when both the planning and execution are complete.}

\begin{figure*}[]
 \centering
 \begin{tabular}{ccc}
 \includegraphics[height=40mm]{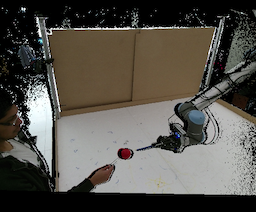} &
 \includegraphics[height=40mm]{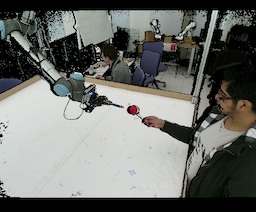} &
 \includegraphics[height=40mm]{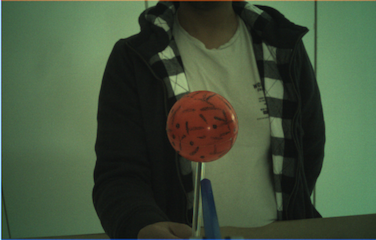} \\
 \includegraphics[height=40mm]{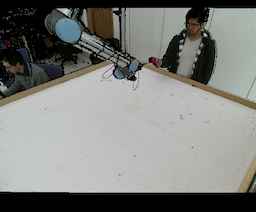} &
 \includegraphics[height=40mm]{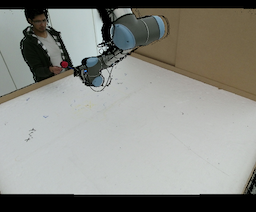} &
 \includegraphics[height=40mm]{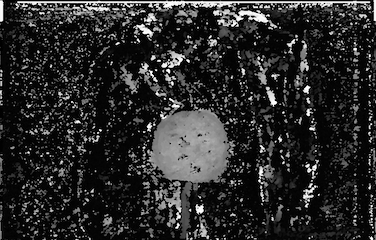} 
 \end{tabular}
 \caption{Input images from 4 Kinect cameras (left and middle) and stereo sensor (right) working in disparity range corresponding to 20-40 cm depth. Data captured during red ball touching experiment (Sec.~\ref{sec:exp-ball}).}
 \label{fig:camera-data}
\end{figure*}

\begin{figure*}[]
 \centering
 \begin{tabular}{ccc}
 \includegraphics[width=0.42\linewidth]{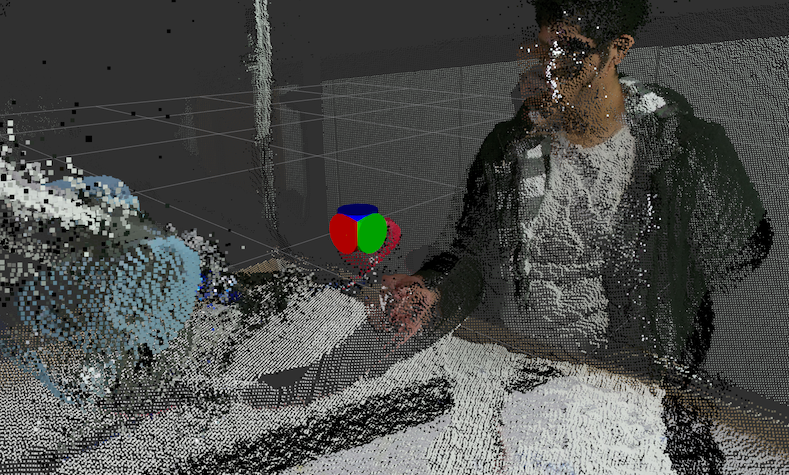} &
 \includegraphics[width=0.4\linewidth]{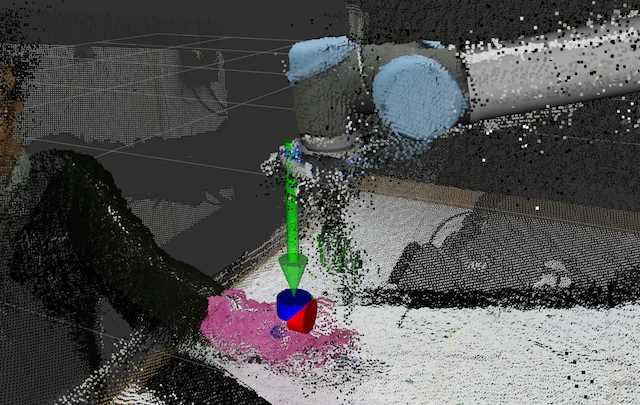} 
 \end{tabular}
 \caption{Point cloud from Kinect sensors combined with tracked pose of the target indicated $\mathbf{X}_t$ as shown in RViz for red ball (left) and hand (right), green arrow is the current goal.}
 \label{fig:point-cloud}
\end{figure*}

\section{Experiments}\label{sec:exp}

This section presents four experiments to demonstrate the proposed method and evaluate its accuracy.
The four Kinect sensors were connected to a workstation (8 cores i7 CPU, GTX1080 GPU), which processed the EtoH data, providing detections at $\sim$5~Hz.
A second identical workstation in the ROS network controlled the UR10 inverted arm, processed data from EinH synchronized stereo sensor ($\sim$10~Hz detections) and hosted the ROS Master controller node.

\begin{figure}[]
\centering
\begin{subfigure}{.22\linewidth}
  \centering
  \includegraphics[height=.7\linewidth]{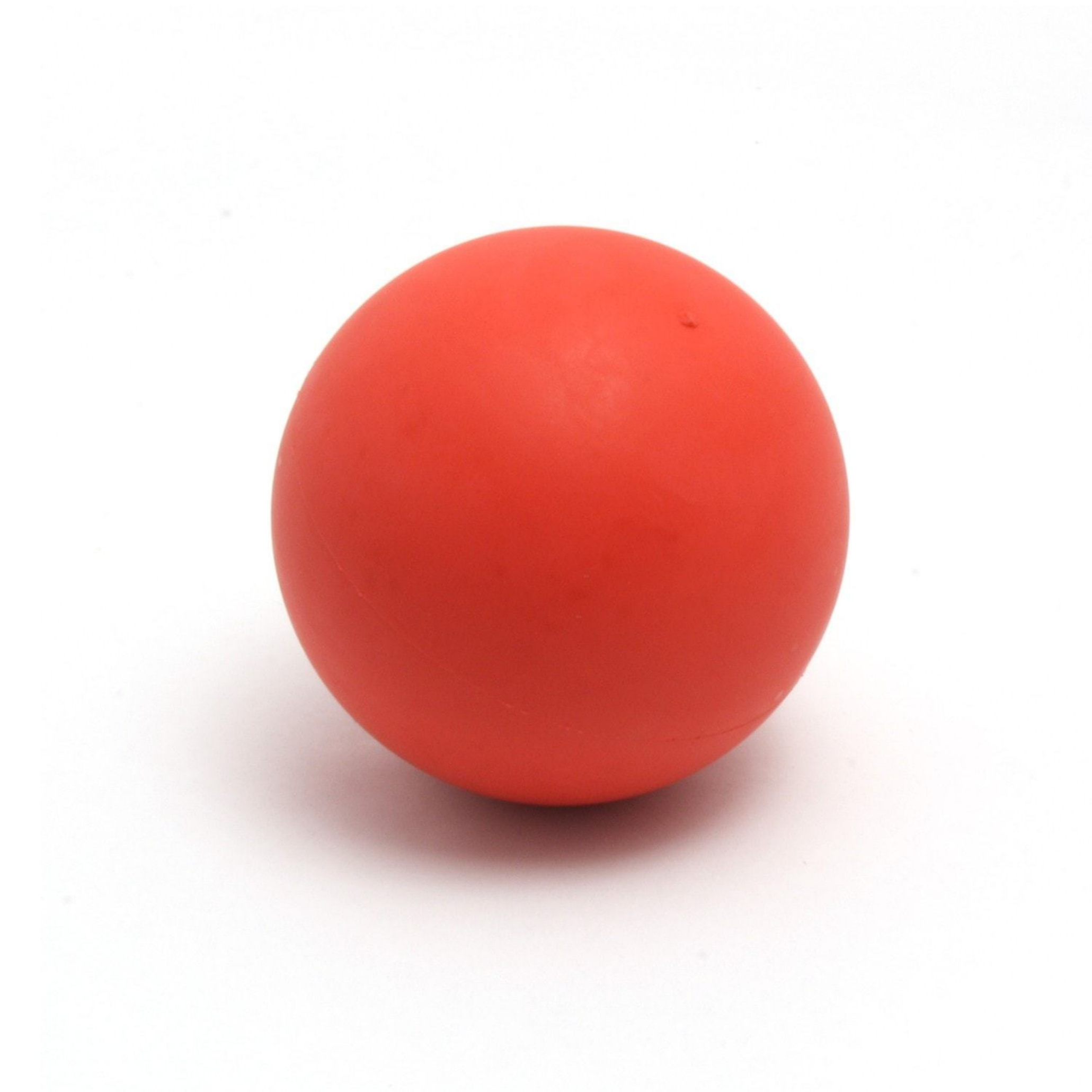}
  \caption{Red ball}
  \label{fig:sfig1}
\end{subfigure}%
\begin{subfigure}{.22\linewidth}
  \centering
  \includegraphics[height=.7\linewidth]{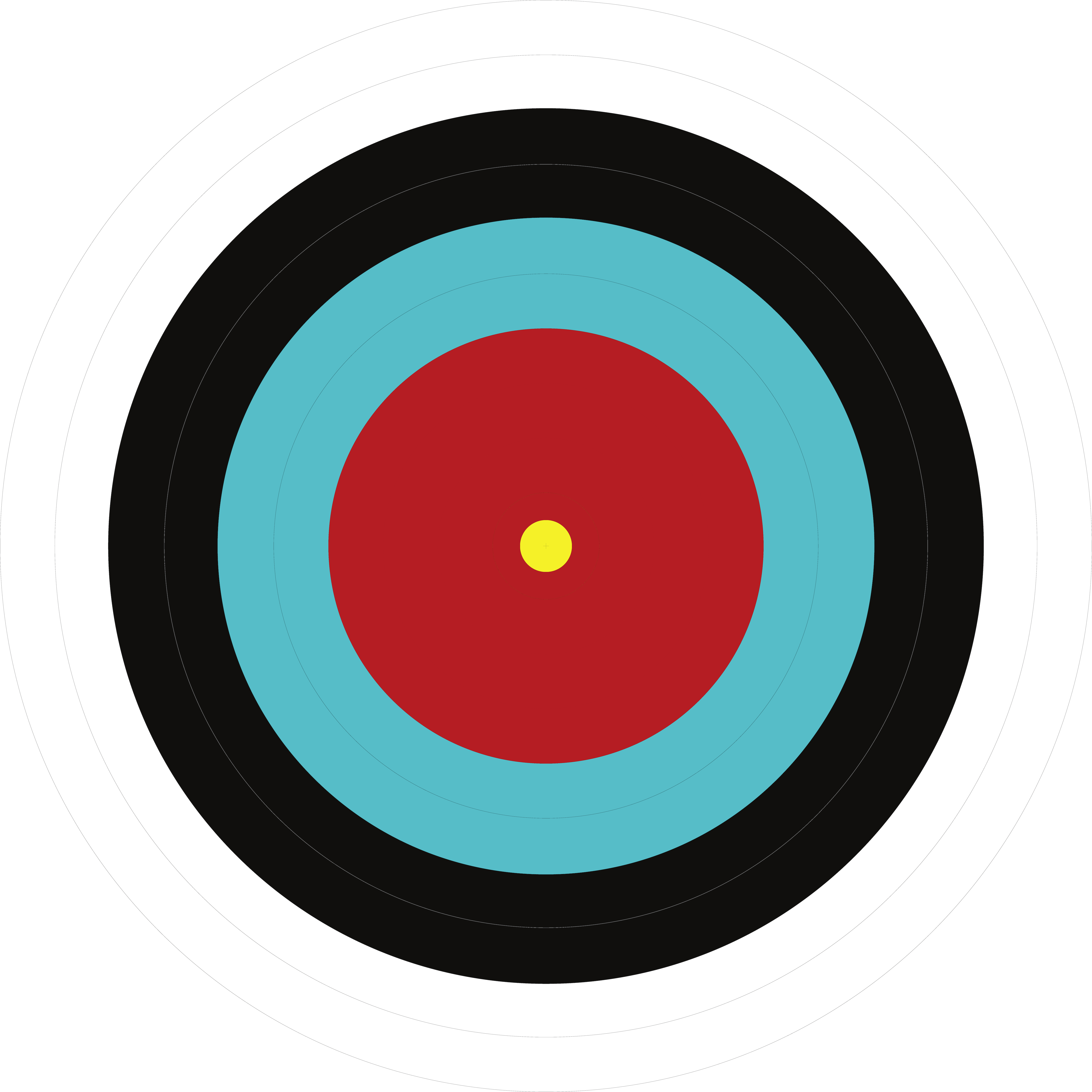}
  \caption{Bulls-eye}
  \label{fig:sfig3}
\end{subfigure}%
\begin{subfigure}{.22\linewidth}
  \centering
  \includegraphics[height=.7\linewidth]{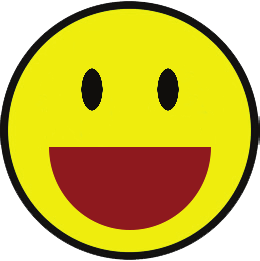}
  \caption{Smiley}
  \label{fig:sfig4}
\end{subfigure}
\begin{subfigure}{.25\linewidth}
  \centering
  \includegraphics[height=.7\linewidth]{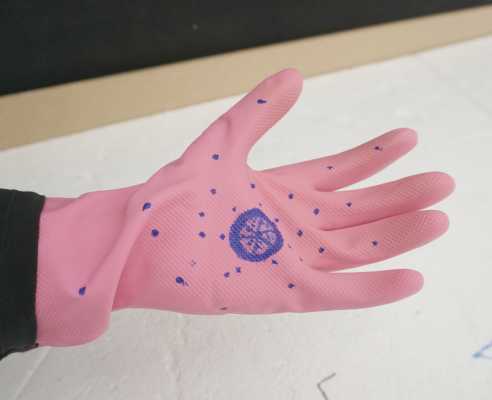}
  \caption{Hand}
  \label{fig:sfig2}
\end{subfigure}%
\caption{Targets used in the experiments.
\label{fig:targets}}
\end{figure}

Examples of the visual servoing input data can be seen in Fig.~\ref{fig:camera-data}. The four images on the left show the four kinect viewpoints covering the workcell. The two images on the right are the color (top right) and disparity (bottom right) images from the EinH stereo sensor.
The targets used in the experiments are seen in Fig.~\ref{fig:targets}.
Fig.~\ref{fig:point-cloud} shows an example of the servoed end effector (colored cube) aligned with the target red ball (slightly visible at the colored cube's edge).

\subsection{Tracking Accuracy}
The dynamic accuracy of both the EinH and EtoH sensors was estimated, with the arm tooltip pose based on joint angle readouts used as the reference.
{\bf EtoH:} The red ball target was attached directly to the tooltip and moved along a predefined trajectory at $\sim$10 cm/s speed. The difference between reference and estimated positions (median distance) was 38 mm.
{\bf EinH:} The bulls-eye target was placed at a known reference position and the arm placed the stereo sensor in front of it within the extent of both the viewing angle (45 deg) and depth range (20-40 cm) of the sensor. Median error distance was 18 mm.

\subsection{Ball Touching} \label{sec:exp-ball}

The red ball target was held by hand and moved randomly by a demonstrator standing on one side of the workcell, while a tip attached to the robot arm endpoint was servoed to touch the ball. For quantitative dynamic evaluation, the ball was moved to 22 waypoints placed at the corners and face centers of a virtual box (100 cm wide, 50 cm high, 50 cm deep), with the demonstrator pausing at each waypoint until servoing converged to its goal.
Every such partial servoing action to a waypoint was successful if the endpoint reached within 5 mm from the surface of the ball. The experiment was performed both in hybrid mode (Kinect+stereo) and Kinect only mode and the median statistics are given in Table~\ref{tab:ball-stats}. The use of EinH in the hybrid mode significantly improves the success rate. The few failures can be attributed to the target estimated at a lower depth than the actual in the stereo sensor, probably due to reflections on the glossy target surface.
The dynamic behavior is best observed in the supplementary video ({\small \url{https://youtu.be/OEiZu0gaP6w}}), which presents all experiments in this section.

\begin{table}[]
\centering
\begin{tabular}{|l|c|c|c|}
\hline
\textit{\textbf{Ball} mode} & \textit{Success rate }& \textit{Time to goal} & \textit{Iterations} \\
\hline
Hybrid & 95 \% & 9.0 s & 11 \\
Kinect only & 68 \% & 10.2 s & 12 \\
\hline
\textit{\textbf{Bulls-eye} mode} & \textit{Accuracy}& \textit{Time to goal} & \textit{Iterations} \\
\hline
Hybrid & 15 mm & 6.4 s & 6 \\
Kinect only & 25 mm & 5.8 s & 8 \\
\hline
\end{tabular}
\caption{Performance of ball touching and bulls-eye aiming scenarios}
\label{tab:ball-stats}
\end{table}

\subsection{Bulls-eye Aiming} 
The bulls-eye target was used to evaluate the accuracy of servoing to static targets.
The servoing was repeated twice for three target locations and four starting endpoint poses, {\it i.e.} 24 total actions. For each servo action which reached the target ($<5$ mm) a point was plotted on the target to mark the endpoint location. The error distance to the target center was subsequently measured, with results summarized in Table~\ref{tab:ball-stats}. 
The EinH sensor improved the accuracy in the final approach stage, where the Kinect system suffered from depth over-smoothing, temporal noise and residual calibration errors. 

\subsection{Head and Straw Docking} 
A potential application of the proposed system is assistance to a disabled person, which can drink from a cup with a straw delivered by the robot to the person's mouth. In our case the person was represented by a 20 cm smiley face (Fig.~\ref{fig:sfig4}) printed on a box and the goal was to insert the straw in the mouth (make contact with the surface).

Similar to the previous experiment the success was evaluated on a set of 24 combinations of start and target poses. 
Flexible straw attached to the cup occasionally deformed on the first contact with the target surface, leading to success of rate 67\%, with 6 iterations or 8.7 s to reach the goal (median). In several failure cases the straw collided with the target box outside the mouth, pushing it away or deforming. 

\subsection{Delivery of Item to Hand} 
\begin{figure}[]
 \centering
 \begin{tabular}{ccc}
 \includegraphics[width=0.4\linewidth]{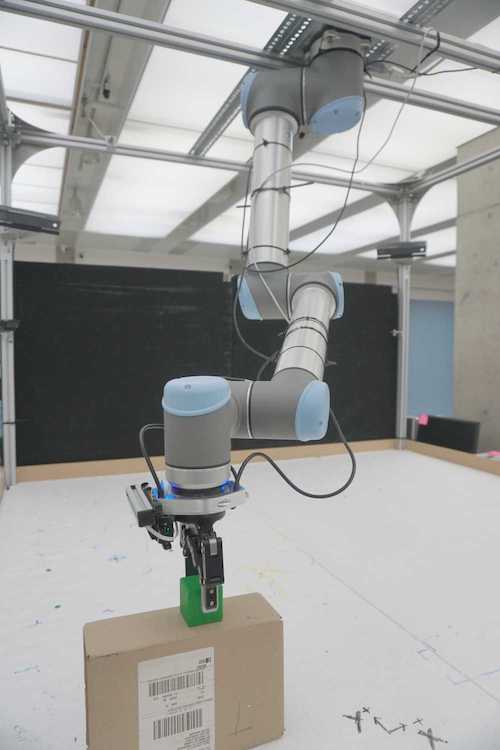} &
 \includegraphics[width=0.4\linewidth]{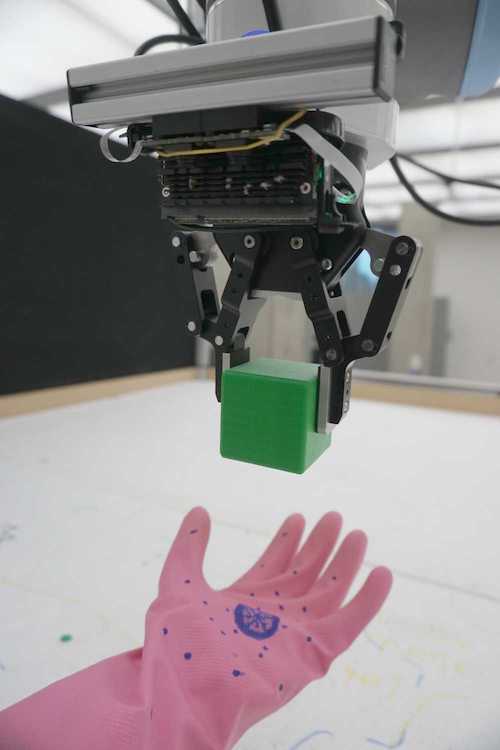} 
 \end{tabular}
 \caption{Delivery of a green cube item from a fixed position (left) to a moving hand (right) using two-finger gripper.}
 \label{fig:hand-pick-drop}
\end{figure}

Another assistance application we include is to pick an item or tool from a fixed location and deliver it to the moving hand (Fig.~\ref{fig:sfig2}) of a person. We control a two-finger Robotiq gripper attached to the robot arm, which releases the item above the palm. For this purpose, the end effector is oriented vertically and EinH camera faces down, as shown in Fig.~\ref{fig:hand-pick-drop}. A pink glove was used for color segmentation of the target in EtoH and a blue palm circle for better localization in EinH.

A set of experiments with the hand moving to 12 different locations repeatedly has shown 75\% success rate of delivering the item to the palm in Hybrid mode, compared to 58\% in Kinect only mode. In some cases depending on the goal approach direction, the palm circle was occluded in the stereo camera view by the item in the gripper, which prevented switching to EinH, resulting in some of the failures.

\section{Conclusions} \label{sec:concl}

We have proposed a system for hybrid visual servoing to moving targets, which achieves a higher success rate and improves accuracy of target reaching when compared to Kinect-only servoing. On the other hand, the increased complexity requires careful calibration of the sensors, which can be difficult to implement.

The experimental evaluation of the proposed approach has exposed several issues. 
1) Although we apply a global calibration method to register the four Kinect sensors, there is alignment error of up to 5 cm error in the far corners of the table, probably due to intrinsic errors in the Kinect sensor. This can lead to extra motion planning cycles to refine the position once moving to the periphery. 
2) The eye-in-hand sensor depth resolution is limited to $\sim$7 mm, which affects targeting error. 
3) The current position based controller limits the servoing cycle to $\sim$1 Hz in practice, when the arm must stop moving before executing a new plan. We are working towards implementing a velocity based controller for the final approach to target, which will allow continuous operation.  



\bibliographystyle{IEEEtran}
\bibliography{IEEEabrv,refiros2018}

\end{document}